\documentclass{article}

\PassOptionsToPackage{numbers, compress}{natbib}

\usepackage[final]{nips_2018}

\usepackage[utf8]{inputenc} 
\usepackage[T1]{fontenc}    
\usepackage{hyperref}       
\usepackage{url}            
\usepackage{booktabs}       
\usepackage{amsfonts}       
\usepackage{nicefrac}       
\usepackage{microtype}      
\usepackage{natbib}
\usepackage{graphicx}

\title{Feed-Forward Neural Networks Need Inductive Bias to Learn Equality Relations}

\author{
  Tillman Weyde \\
  Department of Computer Science \\
  City, University of London\\
  London, United Kigndom \\
  \texttt{t.e.weyde@city.ac.uk} \\
  \And
  Radha Manisha Kopparti \\
  Department of Computer Science \\
  City, University of London\\
  London, United Kingdom \\
  \texttt{radha.kopparti@city.ac.uk} \\
}

\begin{document}

\maketitle

\begin{abstract}
Basic binary relations such as equality and inequality are fundamental to relational data structures. 
Neural networks should learn such relations and generalise to new unseen data. 
We show in this study, however, that this generalisation fails with standard feed-forward networks on binary vectors. 
Even when trained with maximal training data, standard networks do not reliably detect equality. 

We introduce differential rectifier (DR) units that we add to the network in different configurations. 
The DR units create an inductive bias in the networks, so that they do learn to generalise, even from small numbers of examples and we have not found any negative effect of their inclusion in the network. 
Given the fundamental nature of these relations, we hypothesize that feed-forward neural network learning benefits from inductive bias in other relations as well. 
Consequently, the further development of suitable inductive biases will be beneficial to many tasks in relational learning with neural networks. 

 
\end{abstract}

\section{Introduction}
Basic relations such as equality are fundamental to relational data structures. 
One goal of applying neural networks to relational data is that the networks learn to infer these relational structure from data. 
Although equality is typically not learned from data, equality or approximate equality may be embedded as part of other tasks. 
The modelling of equality is clearly in the hypothesis space of feed-forward neural networks (FFNNs) \cite{leshno1993multilayer}, but \cite{GaryMarcus1999,GaryMarcus2001} already highlighted that learning of identity relationships with neural networks may not generalise to unseen data. 
Therefore, we see learning to recognise equality as relevant from a theoretical and practical perspective. 


In this study we test whether feed-forward networks learn equality 
as well as a numeric comparison, thresholded digit sum, and digit reversal of pairs of binary vectors and then generalise this to new data in different settings regarding the task, the amount of data provided, and the depth of the network. 
We find that the recognition of binary relations is not generalised reliably by feed-forward networks. 


To address this problem, we introduce an inductive bias with additional predefined network structures, that we call differential rectifier (DR) units. 
We find  in our experiments that DR units induce reliable perfect generalisation for equality and all other tasks except in digit reversal. 

We see two questions that these results raise:
First, which other relations neural networks do not learn and what that means for more complex tasks. 
Second, what kinds of inductive biases to design and how to implement them. 

The remainder of this paper is organised as follows: Section \ref{sec-related} reviews related literature, Section \ref{sec-task} introduces the task of learning vector equality and our DR units for inductive bias.
Section \ref{sec:exp-res} presents the experimental results and in  Section \ref{sec:conclusions} follow the conclusions of this paper. 


\section{Related work}\label{sec-related}
In relational learning, equality is often not learned from the data, with the exception of the work by \cite{santoro2017simple} who learn to detect equality attributed of objects from images. 
Learning equality could be interesting in the context of constraint learning \cite{de-readt-2018-learning} to learn when equality constraints should be regarded as satisfied. 
Another relevant area is rule learning and application, where soft unification like in \cite{campero2018logical} could be replaced with a learnt model. 

Since neural networks are currently by far the most popular machine learning method, 
it seems of interest whether they can learn equality. 
There have been a number of theoretical contributions showing that feed-forward networks are universal approximators, 
most generally to our knowledge by \cite{leshno1993multilayer}. 
Presumably because of these results there was relatively little interest in the question which functions neural networks can not learn. 
One of the few studies in this direction was undertaken in \citep{GaryMarcus1999} in 1999, where a recurrent neural network failed to distinguish abstract patterns, based on equality relations between sequence elements, although seven-month-old infants showed the ability to distinguish them after a few minutes of exposure.
This was followed by an lively exchange on rule learning by neural networks and in human language acquisition, 
where results by \citep{Elman1999, Altmann2, Shultz1} could not be reproduced by \citep{vilcu2001generalization,vilcu2005two} and 
\cite{shultz2006neural} disputed claims by \citep{vilcu2005two}.
Other approaches, such as \citep{Shastri-1999-spatiotemporal,Dominey,Alhama}, use different network architectures, problem formulations or evaluation methods. 

A more specific problem of learning equality relations was posed in \cite{GaryMarcus2001} by showing that learning of equality on even numbers does not transfer to odd numbers in binary representation.
This relates to the input neuron for the least significant bit not being set to $1$ during training. 
Recently, \cite{mitchell2018extrapolation} addressed this specific problem with different approaches as an example for extrapolation and inductive biases for machine learning in natural language processing. 
However, they did not address the general question of learning equality with neural networks. 

If standard neural networks do not generalise equality relations despite the solution being in their hypothesis space, as we will show for FFNNs below, then the question is how we can enable the learning of solutions that do generalise. 
Inductive biases as a solution can be realised in a number of ways and have been of increased interest recently \cite{hamrick2018relational,snell2017prototypical}.



\section{Equality relation learning}\label{sec-task}
The studies listed above motivated the approach taken here to study a reduced problem outside common contexts such as image analysis or cognitive modelling: whether feed-forward neural networks trained with back-propagation generally have the ability to learn equality relations and generalise to unseen data. 

\begin{figure}[htb]
\centerline{a) $\vcenter{\hbox{\includegraphics[width=3.8cm]
{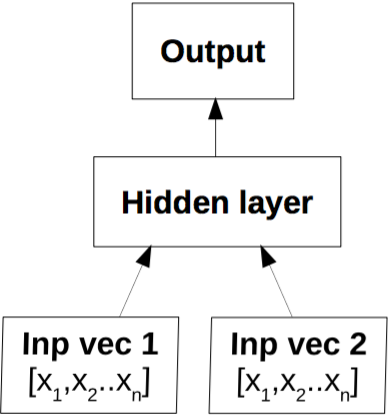}}}$
b)
$\vcenter{\hbox{\includegraphics[width=5.5cm]{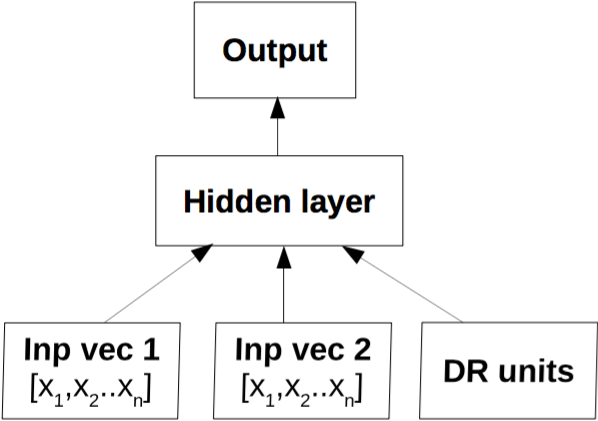}}}$ 
 c) 
 $\vcenter{\hbox{\includegraphics[width=3.8cm]{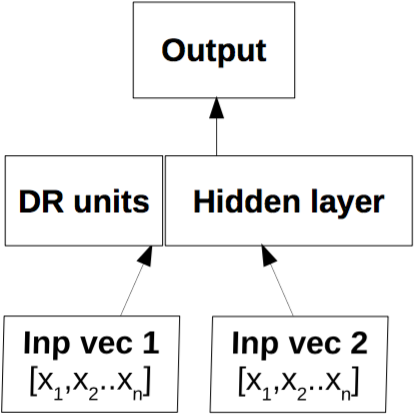}}}$}
 \caption{Network architectures: a) standard feed-forward network, DR integration with b)  early fusion and c) mid fusion. The DR units receive their input in both cases from $vec\hspace{2pt}1$ and $vec\hspace{2pt}2$.}
 \label{fig:nns}
\end{figure}

The general task is to learn the relation between pairs of binary vectors. 
This leads to a binary classification of the pairs according to the equality or otherwise of its element vectors. 
We use a standard FFNN  as sketched in Figure \ref{fig:nns}a). 
This network has $2n$ input neurons, where $n$ is the vector dimensionality. 
The hidden layer has 10 neurons with ReLu activation.
The output layer has two neurons representing the two classes (equal/unequal), which use softmax activation.
The training uses the Adam optimiser\cite{kingma2014adam} with cross-entropy loss. 

The data we train and test the network with is synthetically generated and we vary the type and the distribution of the data in the experiments below. 
We are interested in how many training examples are needed until the network learns to correctly classify pairs of equal vs. unequal vectors. 
This network, like the following ones have been implemented in Python using PyTorch (\url{http://pytorch.org}). 

\paragraph{Inductive bias creation with DR units}
In our model, we use differential rectifier (DR) units that compare input values by calculating the absolute
difference: $f(x,y) = |x-y|$. 
We create one DR unit for every vector dimension with weights from the inputs to the DR units fixed at $1$, 
thus learning the suitable summation weights for the DRs is sufficient for creating a generalisable equality detector.

We use two ways of integrating DR units into the neural networks: \textit{Early Fusion} and \textit{Mid Fusion}.
In Early Fusion, DR units are concatenated to input units \ref{fig:nns}b), and in Mid Fusion they are added to the hidden layer \ref{fig:nns}c). 
In both cases the existing input and hidden units are unchanged. 

\section{Experiments and Results}\label{sec:exp-res}

We performed different sets of experiments using binary vectors for estimating vector equality in relation to vector dimensionality, data size and dataset structure. 
We also use two additional tasks to test the effect of DR units in different contexts.

\paragraph{Effect of network architecture and vector dimensionality}

We generate pairs of random binary vectors with dimensionality $n$ between 2 and 100 as shown in table~\ref{tab:eval2}.
We use all the possible binary vectors to generate equal pairs, i.e. $2^n$ pairs, for $n<10$ and a random selection of 1000 vectors otherwise. 
We also generate the same number of randomly selected unequal vector pairs. 
Then we use stratified sampling to split the data 75:25 into train and test set.
The network is then trained for 20 epochs, which led to convergence in all cases. 
We run 10 simulations for each configuration. 
The average results are shown in Table \ref{tab:eval2}.

We see that the standard FFNNs never fully generalise, and in many cases barely exceed chance level (50\%). 
The early fusion model improves results, but only reaches full performance for 100 dimensions. 
The Mid Fusion reaches perfect test performance in all cases. 

For the plain FFNN, it looks like there is a trend towards better performance at higher dimensionality, but with the observed variation that may be coincidental. 
We did not perform an exhaustive grid search over all hyperparameters, but tested higher numbers of hidden layers (2,3), and larger hidden layers (20,30 neurons) without observing a significant change in the results.
\begin{table}[tb]
 \begin{center}
 \begin{tabular}{|l|r|r|r|}
  \hline
  Vector Dimensions &  Plain FFNN & Early Fusion & Mid Fusion\\ \hline
   n=2  & 52\% &  82\% & 100\%  \\
   n=3  & 55\% &  75\% & 100\%  \\
    n=5  &37\% &  67\% & 100\%  \\
    n=10  &52\% &  75\% & 100\%  \\
    n=30 & 65\% & 75\% & 100\%\\
    n=100  &  75\% & 100\% &  100\%\\
 \hline
 \end{tabular}
\end{center}
 \caption{Accuracy of the different network types on 
 pairs of vectors of different dimensions. The joint train and test data covers all possible equal vector pairs up to 1000, and a random selection where there are more. Only the Mid Fusion architecture reaches reliable equality detection.}
 \label{tab:eval2}
\end{table}

\begin{figure}[tb]
\centerline{\includegraphics[width=8.49cm]
{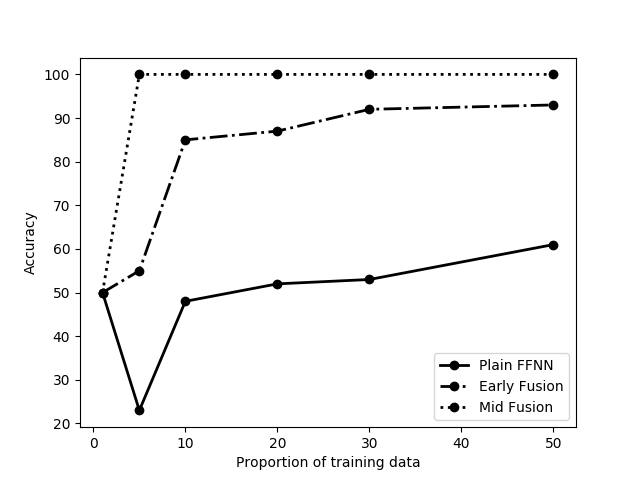}}
\caption{Accuracy of FFNN for 10 dimensional binary vectors after varying the distributions of training data from (1\%-50\%) keeping the testing data fixed}
\label{fig:var}
\end{figure}

\paragraph{Effect of training data size}
We study here how much the performance depends on the training data size.
For this, we vary only the training data size and keep the test set and all other parameters constant. 
We use training data sizes of 1\% to 50\% (in relation to the totally available data as defined above) and the accuracy achieved in various conditions is plotted in Figure~\ref{fig:var}. 
It is worth noting, that the Mid Fusion network reaches 100\% accuracy from 10\% data size on while the FFNN shows only small learning effects. 

\paragraph{Effect of vector coverage}
A possible hypothesis for the results of the FFNN is that the coverage of the vectors in the training set plays a role. 
To share vectors in equal pairs between training and test set would mean to train on the test data, but we created a training set that contains all vectors that appear in the test set in the unequal pairs for $n=10$. 
The results are shown in column a) of Table~\ref{tab:eval3}.  
We also created a training set where each vector appeared as above, but in both position 1 and 2. 
The results are shown in column b) of table~\ref{tab:eval3}.
The results in both cases are similar to those without this additional coverage  in Table~\ref{tab:eval2}.

\paragraph{Other classification tasks} We evaluate here whether the DR units have a negative effect on other learning tasks (using $n=3$). 
We evaluated the networks on the classification by comparing the two vectors in the pair as binary numbers with results shown in column c) of Table~\ref{tab:eval3}. 
We also tested a task that is not a comparison of the two vectors in the pair, by calculating the digit sum. 
We classify by checking if the digit sum is $\geq 3$. 
In both c) and d) we see, that the performance is actually not hindered but helped by the DR units. 
We finally tested the task of recognising digit reversal (swapping least with most significant bits), which DR units are not designed for, as they compare corresponding digits. 
As we can see in column e), DR units do not deliver a perfect solution here, but still lead to somewhat better results than a plain FFNN.  
\begin{table}[tb]
\begin{center}
\begin{tabular}{|l|r|r|r|r|r|}
\hline
 Type  &  a) & b) & c) & d) & e) \\
\hline
1)  Plain FFNN  &  50\% & 52\% &  75\%  &  77\% & 50\% \\
2)   Early Fusion & 75\% & 87\% & 92\%  &  82\% & 55\% \\
3)   Mid Fusion &  100\% & 100\% & 100\% &  100\%  & 58\%\\
\hline
\end{tabular}
\end{center}
\caption{Test set accuracy of FFNN withour and with DR units for different vector coverage (a,b, see text for details) and for classification by c) numeric comparison ($\geq$), d) digit sum $\geq 3$, and e) inversion of digits.}\label{tab:eval3}
\end{table}

\section{Conclusions}\label{sec:conclusions}
In this study we examined the learning behaviour of feed-forward neural networks in vector equality detection and observed that the networks do not generalise well to unseen data. 
We also had similar results in other tasks like numeric inequality and sum of bits of binary vectors. 
We therefore introduced a simple modification to the network with differential rectifier (DR) units and noticed substantial improvements on unseen test data. 
This improvement is largely independent of vector dimension, data size and other parameters. 

The question why standard FFNNs do not learn vector equality relations in a generalisable way is a relevant one, and deserves further theoretical and empirical study. 
It is also important to investigate the design of further measures for creating and controlling inductive biases in neural network learning, as we find that even relatively simple tasks like generalising equality require them. 

\bibliography{nips_2018}
\end{document}